\documentclass[letterpaper, 10 pt, conference]{ieeeconf}  

\usepackage{graphicx}
\usepackage{multirow}
\usepackage{amsmath}
\usepackage[caption=false]{subfig}
\usepackage{cleveref}
\usepackage{amsmath}
\usepackage{mathtools}
\usepackage{afterpage}
\usepackage[table,xcdraw]{xcolor}
\usepackage{graphicx}

\usepackage{stfloats}
\IEEEoverridecommandlockouts                              

\title{\LARGE \bf
AeroVR: Virtual Reality-based Teleoperation with Tactile Feedback \\for Aerial Manipulation
}

\author{Grigoriy A. Yashin, Daria Trinitatova, Ruslan T. Agishev, Roman Ibrahimov, and Dzmitry Tsetserukou
\thanks{The authors are with the Intelligent Space Robotics Laboratory, Space CREI, Skolkovo Institute of Science and Technology, Moscow, Russian Federation.
        {grigory.yashin@skolkovotech.ru, daria.trinitatova@skoltech.ru, ruslan.agishev@skoltech.ru, roman.ibrahimov@skoltech.ru, d.tsetserukou @skoltech.ru}}%
}
\begin{document}
\maketitle
\thispagestyle{empty}
\pagestyle{empty}

\begin{abstract}

Drone application for aerial manipulation is tested in such areas as industrial maintenance, supporting the rescuers in emergencies, and e-commerce. Most of such applications require teleoperation. The operator receives visual feedback from the camera installed on a robot arm or drone. As aerial manipulation requires delicate and precise motion of robot arm, the camera data delay, narrow field of view, and blurred image caused by drone dynamics can lead the UAV to crash. The paper focuses on the development of a novel teleoperation system for aerial manipulation using Virtual Reality (VR). The controlled system consists of UAV with a 4-DoF robotic arm and embedded sensors. VR application presents the digital twin of drone and remote environment to the user  through a head-mounted display (HMD). The operator controls the position of the robotic arm and gripper with VR trackers worn on the arm and tracking glove with vibrotactile feedback. Control data is translated directly from VR to the real robot in real-time. The experimental results showed a stable and robust teleoperation mediated by the VR scene. The proposed system can considerably improve the quality of aerial manipulations.   

\end{abstract}


\section{Introduction}
The technologies aimed at aerial manipulation by unmanned aerial vehicle (UAV) draw deep attention from companies of e-commerce (e.g., Amazon, Walmart, etc.). In many applications, such as industrial maintenance, structure inspection, it is required to carry out aerial manipulation with the objects. For example, picking and delivering the parcel, inspection of the pipelines and bridges with the sensors, holding a heavy object with the swarm of drones, etc. In such situations, it is required to equip a drone with a dexterous robotic manipulator.
Nowadays, several research teams are working on the development of UAV with a robotic arm. The control system, dynamics, and stabilization of multirotor with manipulators are studied in \cite{korpela2013dynamic}, \cite{jimenez2013control}, \cite{kim2013aerial}. In the paper of T. W. Danko and P. Y. Oh \cite{danko2014toward}, the manipulator is visually servoed using an eye-in-hand camera. Flying robot which is capable of grasping objects in flight is presented in \cite{thomas2014toward}. In \cite{sarkisov2019development}, the cable-suspended aerial manipulator is developed for safe aerial manipulation in a complex environment. A. Suarez et al. \cite{suarez2017lightweight} developed the flying robot with the variable parameter integral backstepping controller for industrial maintenance. Japanese company PRODRONE demonstrated multirotor with two manipulators, which can lift objects from the ground and land on the railing by using two manipulators \cite{prodrone}.

\begin{figure}[t]
\centering
\includegraphics[width=0.9\linewidth]{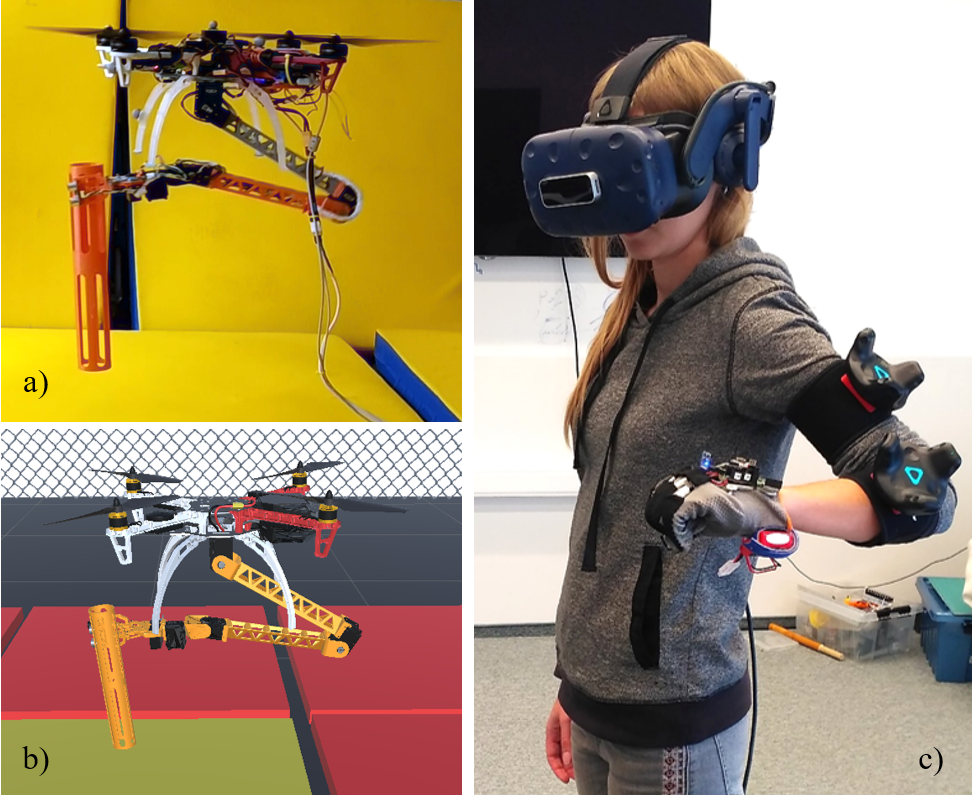}
\caption{The flight test of the aerial manipulation. a) Robot during teleoperation. b) VR visualization. c) Operator with HMD and motion capturing system.} 
\label{fig:rscheme} 
\vspace{-1.5em}
\end{figure}

\par
It is worth to mention that the delivery of objects is of interest not only for commercial purposes but also for life-saving, security, posting, etc. Industrial maintenance and assistance in rescue operations in dilapidated buildings are the two most important tasks for which flying robots capable of performing aerial manipulation are of the particular interest. For this, the robot should carry out specific tasks with high accuracy that could not be performed in autonomous mode. This can be achieved with teleoperation. Additionally, in this case, the robot and the operator can be located in the remote places, which is important for hazardous conditions. According to J. J. Roldan et al. \cite{roldan2019multi}, the integration of the robots with virtual reality interface can be the most convenient way to remotely control the robots. A solution for inspecting industrial facilities via teleoperation is proposed in \cite{suarez2018lightweight}. A. Suarez et al. designed a hexarotor platform equipped with a 2-DOF compliant joint arm which is controlled by a wearable exoskeleton interface via visual feedback. Despite the proposed concepts and systems, to our knowledge, the experiments of teleoperation of aerial manipulation have never been demonstrated in real conditions.
\par
The system proposed in this paper (Fig. \ref{fig:rscheme}) is aimed at accurate object manipulation by the UAV equipped with a robotic arm in a remote environment when there is no direct visual contact between a human operator and drone. This goal is achieved by specially developed human-robot interaction strategy. In our case, the robotic arm attached to the quadrotor is capable of reproducing human hand movements. Such intuitive control allows an operator to manipulate remote objects successfully without any preliminary training. In addition to this, the system provides an operator with the tactile feedback, informing whether the target object is grasped. To expand human capabilities to interact with objects of interests in remote or occluded environment, the robot with manipulator as well as the objects are also simulated in virtual reality. Their positions and displacements are transmitted to the VR environment in real-time. This enhances an operator perceptional capabilities about the hardly reachable or remote environment.

\section{System Architecture}
Our solution for the teleoperation of the 4-DoF force-sensing manipulator attached to the flying robot (Fig. \ref{fig:rscheme} (a)) is a virtual reality system that consists of VR application (Fig. \ref{fig:rscheme} (b)), HMD, HTC VIVE trackers (trackers) and a glove (Fig. \ref{fig:rscheme} (c)). Trackers and glove transfer the position and the orientation of the operator's hand to the VR application designed in the Unity platform, which sends target angles to the robot manipulator. The glove controls the closing of the grip and delivers feedback to the operator via vibration motors when the gripper bars interact with an object. For verification of the whole developed system, we tested the VR teleoperation of aerial manipulation using a VICON motion capture system (mocap), which provides localization data of the robot and object.

\subsection{Robot Design}

A payload of more than 1 kg, long continuous flight time, and size are the main UAV requirements. Taking into account these limitations, we assembled a quadrotor based on the frame DJI Flame Wheel ARF KIT F450 with propulsion system DJI E600, and Cube Flight Controller (based on Pixhawk 2.1). The robot is connected to a stationary power supply via cable to eliminate the need for a changing battery during long flight tests.

\begin{figure} [h]
\centering
\includegraphics[width=6.6 cm]{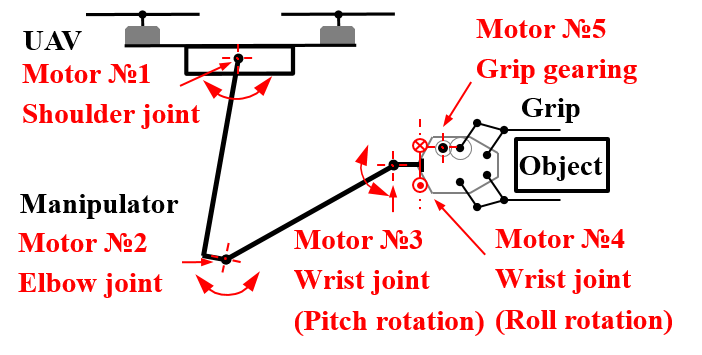}
\caption{Layout of the manipulator servo motors.}
  \label{fig:Manipulator0}
  \vspace{-0.5em}
\end{figure}

\par
4-DoF manipulator (Fig. \ref{fig:Manipulator0}) consists of three motors in articulated joints (Dynamixel servomotors MX-106T, MX-64T and AX-12 in the shoulder, elbow and wrist joints respectively), two links, one servomotor for the grip rotation (Dynamixel AX-12) and the grip with 1-DoF (Futuba S9156). The grip design is based on the four-bar linkage mechanism. The technical characteristics of the developed robot are presented in Table \ref{table_1}.

\begin{table}[h]
\centering
\caption{Technical Characteristics of Flying Robot}
 \label{table_1}
\begin{tabular}{ |p{6.5cm}|p{1.2cm}|}
 \hline
 \cellcolor[RGB]{240,240,240}{Recommended load of motors DJI E600} & 600 g/axis\\
 \hline
 \cellcolor[RGB]{240,240,240}Weight of flying robot  & 2080 g \\
  \hline
\cellcolor[RGB]{240,240,240}Manipulator weight (with electronics)  & 918 g  \\
  \hline
 \cellcolor[RGB]{240,240,240}Maximum robot payload &400 g\\
  \hline
 \cellcolor[RGB]{240,240,240}Manipulator length    &740 mm \\
  \hline
 \cellcolor[RGB]{240,240,240}Distance between the edges of the propeller and end of the grip&   400 mm \\
  \hline
 \cellcolor[RGB]{240,240,240}Number of the manipulator DoF& 4\\
 \hline
\end{tabular}
\vspace{-1.5em}
\end{table}

\par
\begin{figure}[h]
\centering
\includegraphics[width=0.9\linewidth]{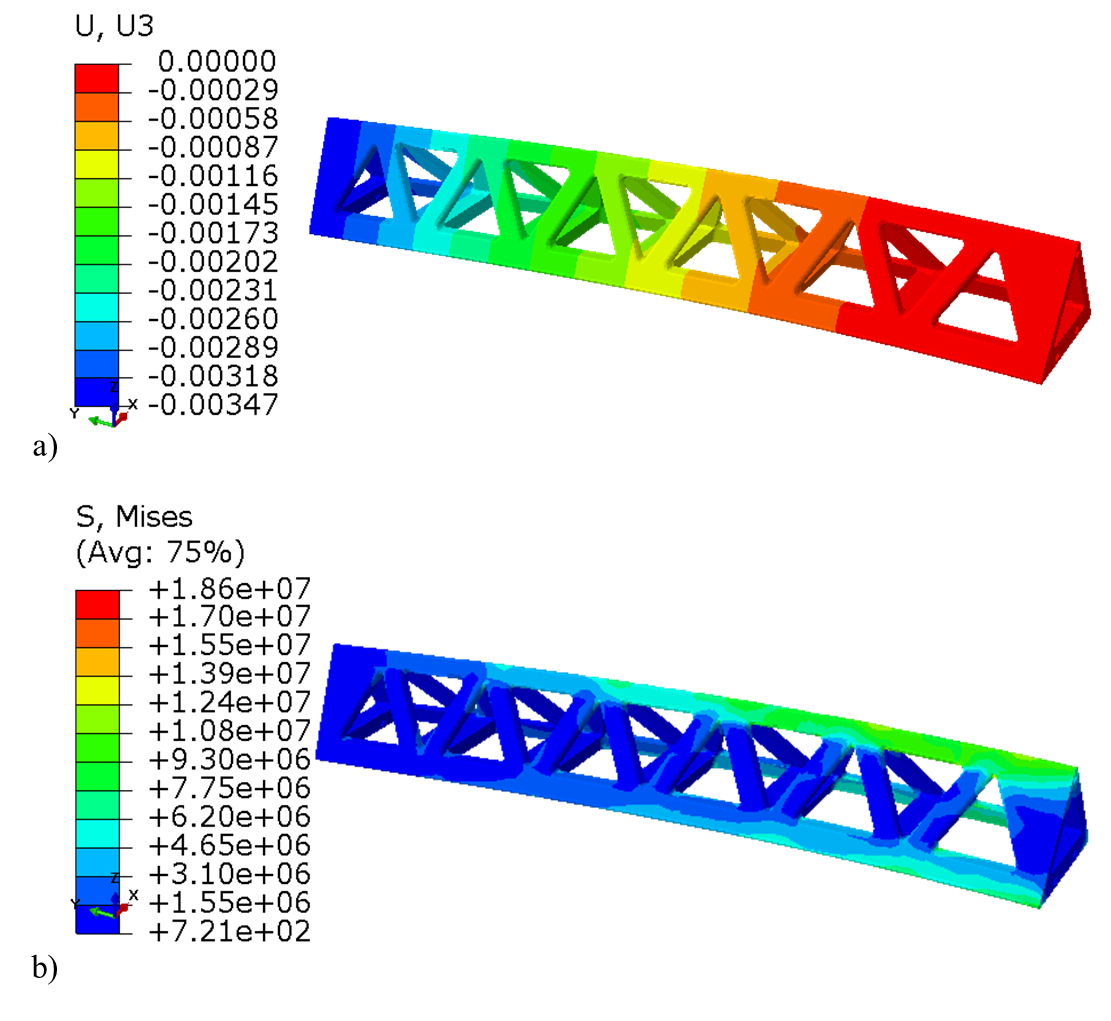}
\caption{The distributions of displacement along Z-axis (a) and stresses (b) in manipulator link.} 
\label{fig:FEMlink} 
\vspace{-1 em}
\end{figure}

 
Manipulator links have a truss structure to achieve a light and rigid construction. The links are 3D-printed from PLA material. The stress-strain analysis of links was carried out in FE Software Abaqus 6.14 to select the most optimal design in terms of minimum weight and maximum rigidity. For the finite element method, the robot link was considered as cantilever beam loaded by uniform gravity force and torque of 5.3 N$\cdot$m at the opposite end (while holding an object weighing 400 grams in the extended position of the manipulator). The distributions of displacement along Z-axis and stresses in the manipulator link are provided in Fig. \ref{fig:FEMlink}(a) and Fig. \ref{fig:FEMlink}(b), respectively. The displacement along Z-axis is less than 3.5 mm at maximum payload, the deflection angle is less than 1 degree. Thus, the manipulator has a robust structure while manipulating an object.
 
\subsection{VR System}
The VR setup includes HTC Vive Pro base stations, HMD, and Vive trackers attached to the arm for tracking the user's hand motion in VR. The first tracker is mounted on the shoulder to control the rotation of the shoulder joint of the manipulator. The second tracker is fastened around the elbow joint and aimed to control the elbow joint rotation of the manipulator. Experimentally we defined the optimal position of trackers, which provides the largest work area of the elbow and shoulder joint angles of the operator's arm.

\subsection{System of Interaction with the Target Object }
The robotic arm is equipped with two force sensors FSR 400 mounted on the gripper bars. These sensors detect the contact between the gripper and the object and determine the force with which the bars of the gripper hold the cargo. 
\par
\begin{figure} [h]
\centering
\includegraphics[width=8.0 cm]{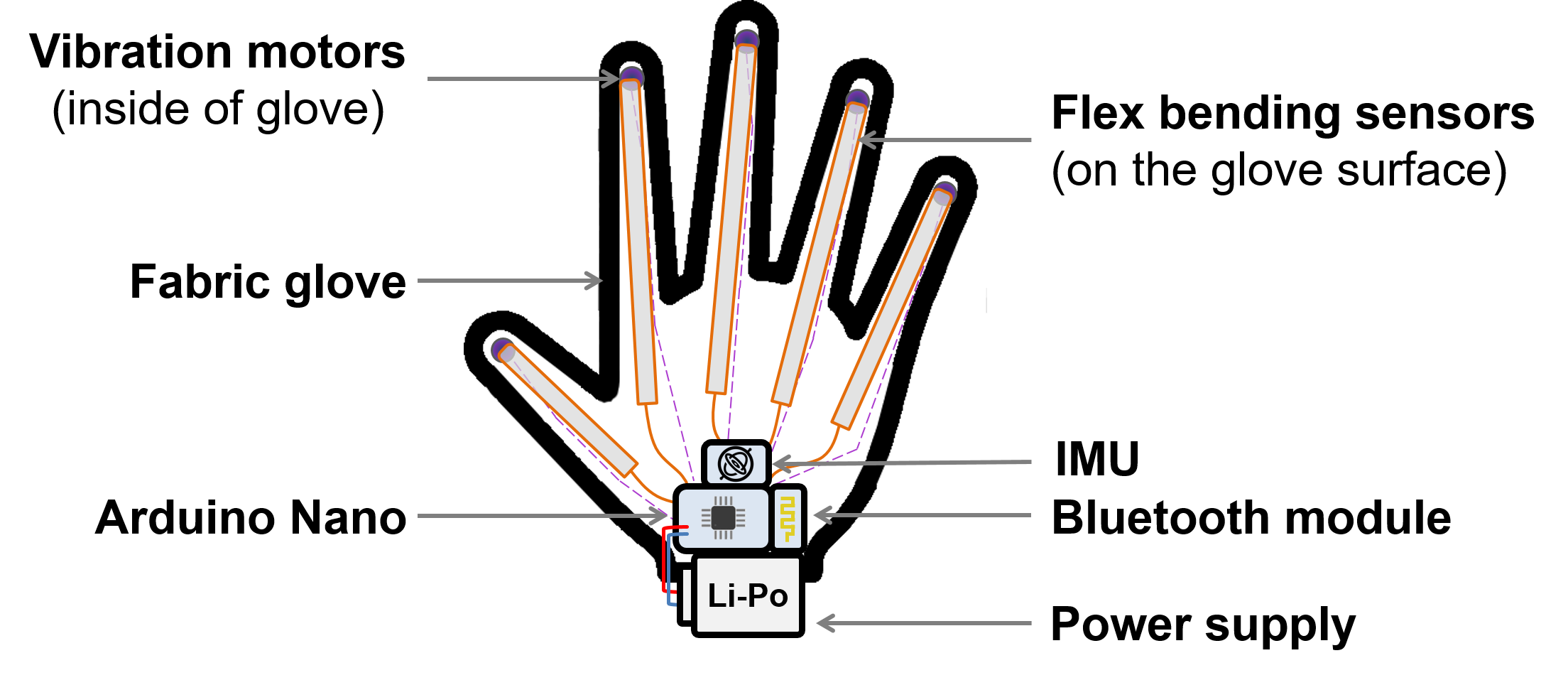}
\caption{The scheme of the smart glove.}
  \label{fig:Glove}
  \vspace{-1.0 em}
\end{figure}

The design of the tracking glove was inspired by the glove of T.K. Chan and et al. \cite{chan2018robust}. They tested the glove equipped with inertial measurement unit (IMU) and flex sensors. Our glove for grip control consists of the fabric glove, 5 embedded flex sensors, coin vibration motors, battery, and control electronics (Fig. \ref{fig:Glove}). Spectra symbol flex sensors 4.5 measure flexion angles of fingers that directly controls the opening/closing of the gripper. IMU sensor AltIMU 10 v4 determines the rotation of the hand in space (angles of the wrist joint). We sewed 5 vibration motors into the glove at the fingertip areas that allows transferring the feedback about the interaction of the manipulator with the grasped object.  The use of the vibration motors on the fingertips to get feedback from the micro UAVs was proposed in the paper \cite{labazanova2018swarmglove}. Control of sensors and vibration motors is performed via Arduino Nano. The data exchange between the microcontroller and computer is carried out by Bluetooth module HC-05. The assembled electrical circuit is powered by a 7.5V Li-Po battery via DC-DC converter MP1584EN.

\section{Control System of the Robot}
\subsection{Manipulator Kinematics}

\begin{figure}[b]
\centering
\includegraphics[width=0.62\linewidth]{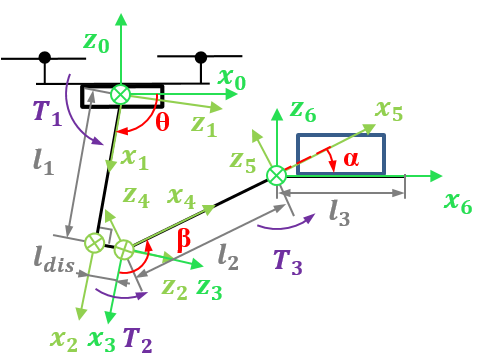}
\caption{Scheme of the manipulator.} 
\label{fig:Sch}
\vspace{-1.0 em}
\end{figure}

We used inverse kinematics to calculate the desired position of the gripper and the possible geometric limitations of the manipulator. For VR teleoperation mode, we applied forward kinematics. At the calculation of the kinematics, we do not take into account the rotational degree of freedom of the grip since we are transmitting the desired angle directly to the servomotor controlling this DoF. Thus, we consider the kinematics of 3-DoF planar manipulator. Fig. \ref{fig:Sch} shows the scheme of the robotic arm. The abbreviations are as follows: $\theta$, $\beta$, and $\alpha$ are the joint angles; $l_1$ and $l_2$ are the link lengths; $l_3$ is the position of the grip end in relation to the wrist joint; $(x_i, z_i)$ coordinate system (CoS) is attached to $i$-th CoS transformation; $l_{dis}$ is the transition from CoS ($x_2, z_2$) to CoS ($x_3, z_3$) due to the shape of link 1; $T_1$, $T_2$, and $T_3$ are torques in shoulder, elbow and wrist joints, respectively.

\subsubsection{Inverse Kinematics}
For the 3-DoF planar manipulator, we used the following transition matrix, which corresponds to the transformation from the CoS ($x_5, z_5$) to drone coordinates (${x_0, z_0}$):

\begin{equation}
    ^B_pT = T^0_5 = 
    \begin{bmatrix} 
\cos(\phi) & -\sin(\phi) & 0 & x \\
\sin(\phi) & \cos(\phi) & 0 & z \\
0 & 0 & 1 & 0 \\
0 & 0 & 0 & 1 
\end{bmatrix}
\label{eq:matr} 
\end{equation}
\par  

where 
$\Bigg\{\begin{matrix} 
\phi= \alpha -\beta + \theta \\ 
x = l_1 \cdot \cos(\theta) + l_2 \cdot \cos(\beta -\theta) - l_{dis} \cdot \sin(\theta) \\
z = l_1 \cdot \sin(\theta) -l_2\cdot \sin(\beta -\theta)+l_{dis} \cdot\cos(\theta)
\end{matrix}$

\quad
\newline
\par Since the system is underconstrained, we decided to impose an additional limitation for $\alpha$ to keep the object parallel to the ground. This condition is supposed to provide the stability of the grabbed object. Firstly, we calculate $\beta$ from (\ref{eq:matr}) using MATLAB function \textit{solve} for symbolic computation. Secondly, using systems of equations (\ref{eq:matr}) and (\ref{eq:syst}) \cite{craig2009introduction} we calculate $\theta$ and $\alpha$:
\newline
\begin{equation}
\Bigg\{\begin{matrix} 
x= k_1 \cdot \cos(\theta) - k_2 \cdot \sin(\theta) \\
z = k_1 \cdot \sin(\theta) + k_2 \cdot \cos(\theta) \\
k_1 = l_2 \cdot \cos(\beta) +l_1 \\
k_2 = -l_2 \cdot \sin(\beta) + l_d
\end{matrix}
\quad
\label{eq:syst}
\end{equation}
\begin{equation}
    \Bigg\{\begin{matrix} 
\theta = -(\arctan(x,y)-\arctan(k_1, k_2)) \\
\alpha = \beta -\theta
\end{matrix}
\quad
\end{equation}

\subsubsection{Forward Kinematics}
We used forward kinematics to display the position of manipulator components in Unity. To do this, it is necessary to write the transformation matrix for each joint of the manipulator via rotation and translation matrices from the drone CoS ($x_0, z_0$) to each joint CoS. Thus, we got the following equations for position of elbow, wrist joints (\ref{eq:elb}, \ref{eq:wr}) and the grip end (\ref{eq:gr}):

    \begin{equation}
    \begin{matrix}
    x_e=x_s+l_1\cdot\cos(\theta)-l_{dis}\cdot\sin(\theta)\\
    z_e=z_s+l_1\cdot\sin(\theta)+l_{dis}\cdot\cos(\theta)
\end{matrix}
\label{eq:elb} 
\end{equation}
\begin{equation}
    \begin{matrix}
    x_w=x_e+l_2\cdot\cos(\beta-\theta)\\
    z_w=z_e-l_2\cdot\sin(\beta-\theta)
    \end{matrix}
\label{eq:wr} 
\end{equation}
\begin{equation}
 \begin{matrix}
x_g=x_w+l_2\cdot\cos(\alpha-\beta+\theta)\\
z_g=z_w+l_2\cdot\sin(\alpha-\beta+\theta)
\end{matrix}
\label{eq:gr}
\end{equation}
where $x_s$, $z_s$ are the coordinates of the shoulder joint that corresponds to the current position of the UAV.

\par 
\subsection{GUI to Monitor the Manipulator State}
\begin{figure*}[t]
\centering
\includegraphics[width=17.2 cm]{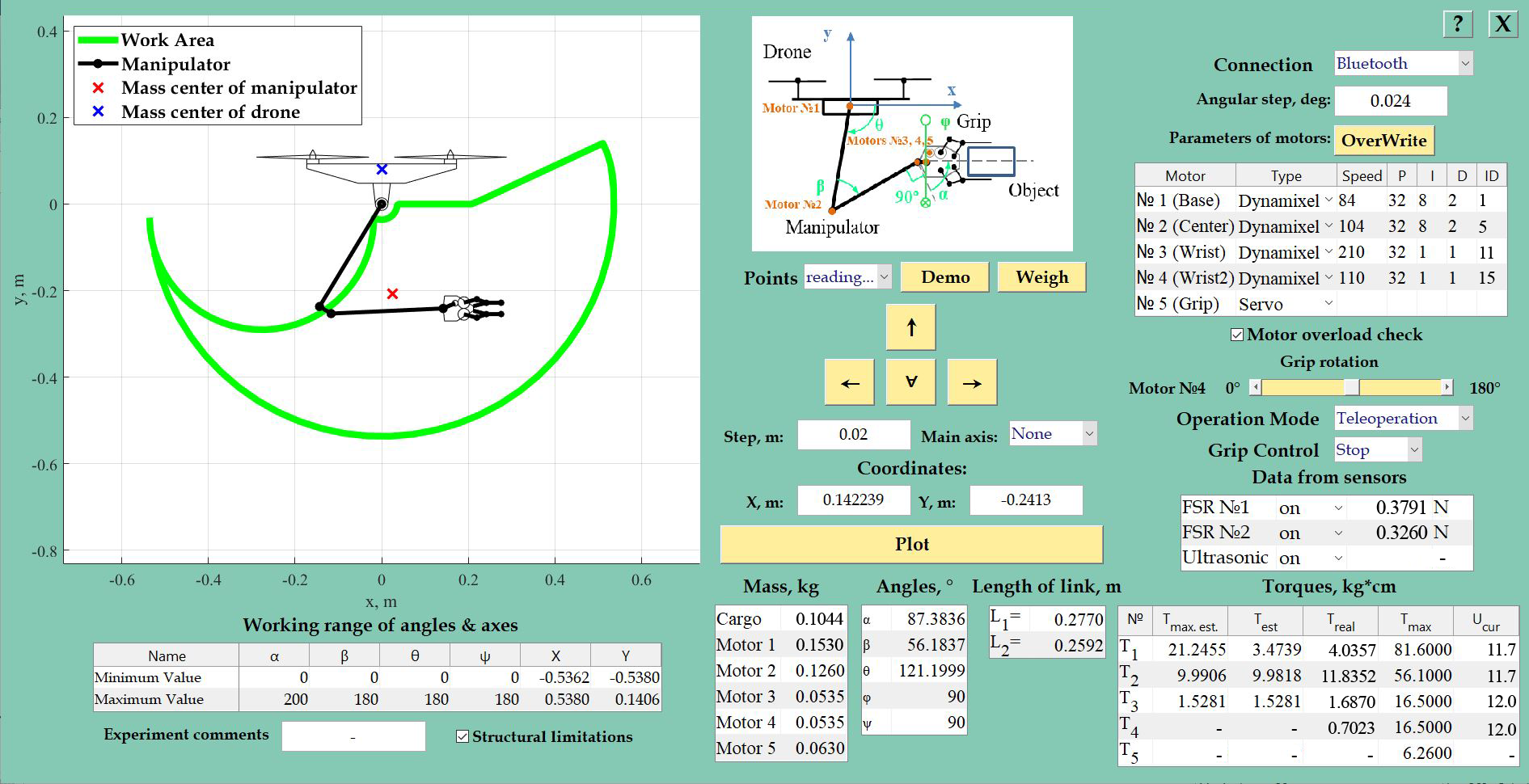}
\caption{GUI to monitor the manipulator state.} 
\label{fig:GUIp}
\vspace{-1.0em}
\end{figure*}

The connection of motors and sensors to MATLAB is performed via board OpenCM 9.04. Wireless control of the robotic arm is carried out by Bluetooth module HC-05. Debugging of the control algorithm and monitoring the state of the robot is realized by the developed GUI interface in MATLAB (see Fig. \ref{fig:GUIp}). Using the GUI, we can conveniently configure the robot and observe the real-time manipulator behavior using a computer. In addition, we can send a command to turn on (off) the manipulator power supply using the transistor switch. It is necessary if the behavior of UAV will become unstable.

\par 
Depending on the manipulator components, GUI provides information about the work area of the manipulator and the required torque of servomotors. It also displays information about the current technical parameters of servomotors (voltage, speed, PID coefficients) and sensor data. The user can change the following manipulator settings:
\begin{itemize}
  \item Control modes of motors depending on the controllers: Arduino UNO, USB2Dynamixel or OpenCM9.04.
  \item Work area via changing the acceptable value of joint angles.
  \item Lengths of manipulator links.
  \item Types of the manipulator motors (ordinary servomotors or Dynamixel smart servomotors). 
  \item Teleoperation and manual manipulator control modes.
\end{itemize}
\par In manual mode, the operator can change the position of the gripper by writing desired X and Y coordinates or pushing on the arrow keys (forward, back, up or down). The step size is set separately for the usage of the arrow keys. Teleoperation mode switches the GUI to the mode of displaying and recording the experimental data, while control of the manipulator is transferred to Unity and the operator.

\subsection{The UAV Localization}
The external position estimation system is used to localize the UAV during the flight. In order to get the high-quality tracking of the quadrotor and a target object during the experiments, we use mocap with 12 cameras (Vantage V5) covering 5 m x 5 m x 5 m space. Robot Operating System (ROS) Kinetic framework is used to run the development software and mavros ROS stack for drone control and ground station communication. The position and attitude update rate for the quadrotor and the target object is 100 Hz. The estimated poses of the tracked robot and object are sent to the computer running Unity 3D software for VR simulation. While the main goal of the experiment is to manipulate an object in virtual and real environments at the same time, the quadrotor is commanded to give in place holding its position close to the object (in this position, the object is located within the robotic arm's manipulation area). During the aerial manipulation, the robot position is corrected by a UAV operator in order to prevent high oscillations around the desired location of the robot.

\section{VR Control System}

\par
\begin{figure} [h!]
  \centering
  \includegraphics[width=0.9\linewidth]{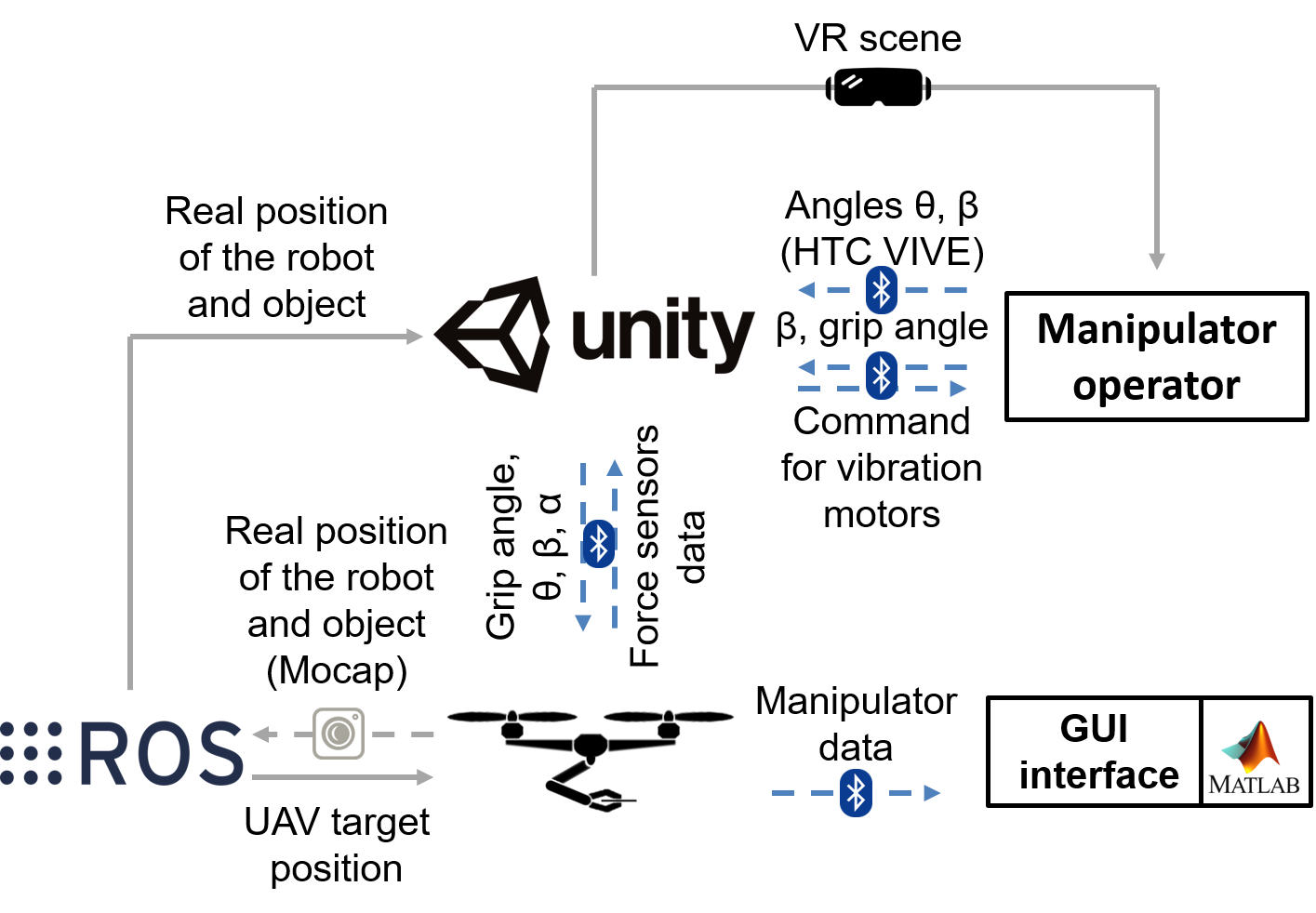}
  \caption{The scheme of communication of the VR-based teleoperation.}
  \label{fig:Scheme}
  \vspace{-1.0em}
\end{figure}

 Fig. \ref{fig:Scheme} shows the scheme of communication among mocap, ROS, Unity, the flying robot, and the operator.
Unity application is the core part of this system, which receives the data from the operator and the robot through a Bluetooth connection and from ROS through a cable. Also, Unity sends the necessary commands to the devices in accordance with predefined scripts. To provide the user with a visual control, the operator wears HMD to experience VR application, which includes a simplified model of the room along with a flying robot. Mocap cameras transfer positioning and rotation tracking data of each object to the Unity 3D engine. The connection between Mocap and VR application is based on the method proposed in \cite{roman2019}. In addition, data from the manipulator are transmitted to the GUI.

\par
\begin{figure} [h]
  \centering
  \includegraphics[width=0.78\linewidth]{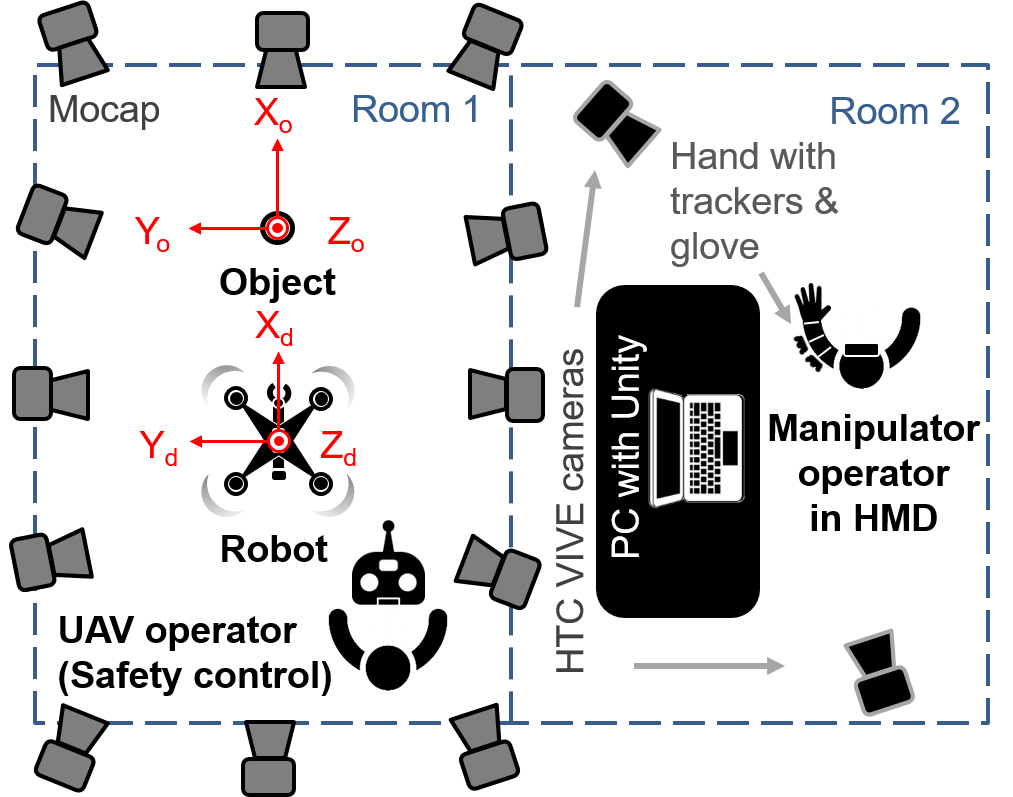}
  \caption{Scheme of the rooms for VR-based teleoperation experiment.}
  \label{fig:Room}
  \vspace{-1.0em}
\end{figure}

 The work space for testing the teleoperation system consists of two rooms (Fig. \ref{fig:Room}). The first room, where the aerial manipulation is carried out, is equipped with mocap cameras. In addition, the UAV operator with a radio remote control is present in this room to eliminate unforeseen high oscillations around the target position. In the second room, we installed HTC VIVE equipment and base station (PC with Unity). The operator can freely move his or her hand on which the trackers and glove are worn. Transfer data about the relative position of the robot and the object in VR from mocap allows the operator to accurately navigate robotic arm in the current situation which corresponds to the real context in the room where the flight experiment is carried out. 

\section{Experiments }
Experiments have been conducted in two stages: the flight test with GUI manual teleoperation and VR-based teleoperation in stationary and flight modes.

\subsection{Teleoperation Tests using GUI}
\begin{figure} [t]
  \centering
  \includegraphics[width=0.88\linewidth]{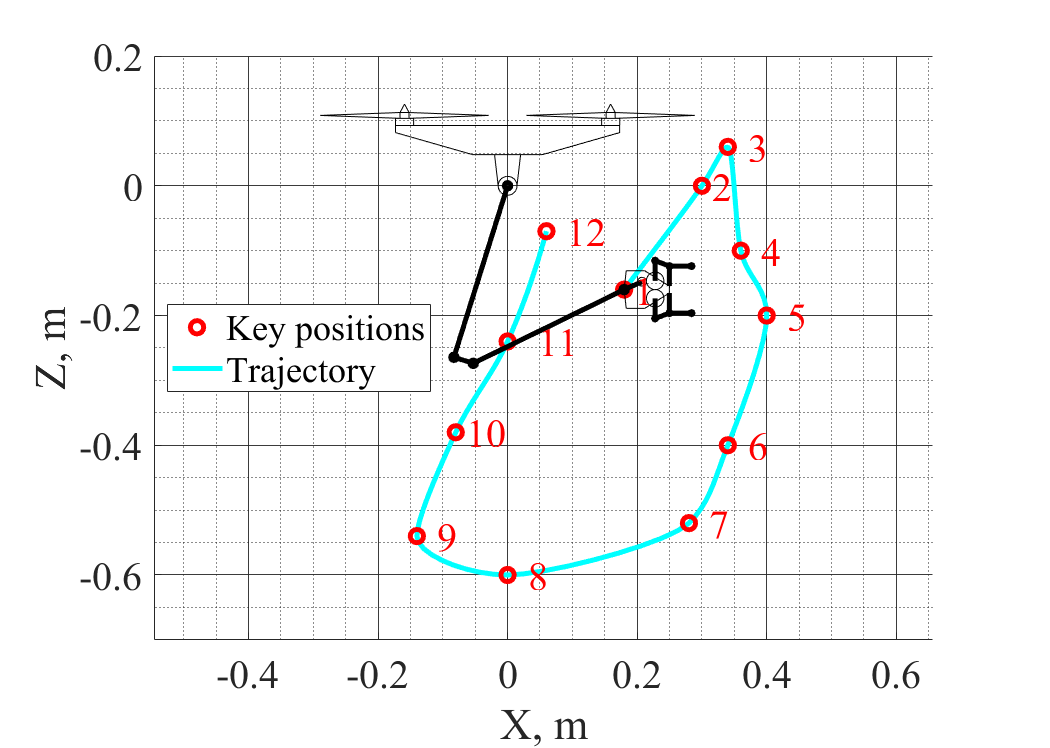}
  \caption{The trajectory of the wrist joint position during experiment.}
  \label{fig:Traj}
  \vspace{-1.0em}
\end{figure}

\begin{figure}[h]
\centering
\includegraphics[width=1\linewidth]{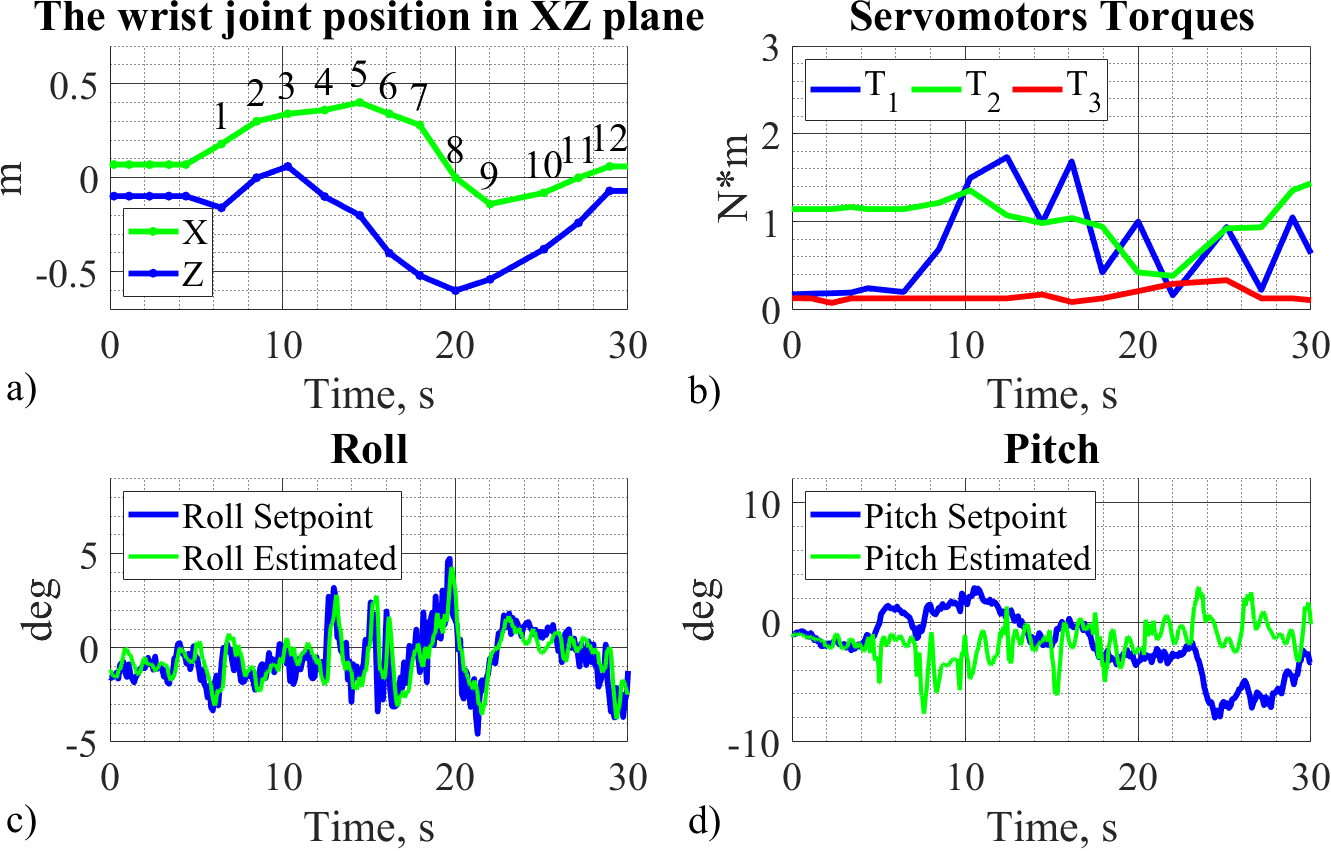}
\caption{Dependences of coordinates (a) and motor torques (b) of the manipulator, roll (c) and pitch (d) angles from the experiment time for the flight test with the GUI teleoperation.} 
\label{fig:test1} 
\vspace{-1.0em}
\end{figure}

In the first trials, we tested the robot operation in flight using a GUI control of the manipulator. We moved the manipulator along the specified path (Fig. \ref{fig:Traj}, Fig. \ref{fig:test1}(a)) which includes a command to drop the object in point No. 9. A wooden cube weighing 77 grams was used as an object for this experiment.

\par Fig. \ref{fig:test1}(c) shows that the maximum deviation of the roll angle is of 5.22 degrees (standard deviation is of 0.99 degrees). The fluctuations in the roll direction in the range of points 3 to 9 arise due to the fast manipulator movement with a considerable distance. The standard  deviation of pitch angle is of 3.51 degrees, the maximum pitch error of 9.38 degrees (see Fig. \ref{fig:test1}(d)) occurs when the manipulator is moved along one of the coordinates about 0.2 m, which is accompanied by a significant change of the servomotor torque in the elbow joint.

\subsection{VR Teleoperation Tests}
\par For the VR teleoperation experiment we chose a plastic cylinder weighing 105 grams as the object to be grasped. During stationary testing of the teleoperation system,  we defined that the position of the trackers on the shoulder and forearm influences the possibility of trackers' circular offset on the hand, which can decrease the work area of $\theta$ and $\beta$. After that, we calibrated delays of control codes for the manipulator, glove, and Unity. Long delay or non-synchronous repetition of the movement of the operator’s hand by the manipulator can cause unsuccessful aerial manipulation with the object.
\par
\begin{figure}[!ht]
\centering
\includegraphics[width=1\linewidth]{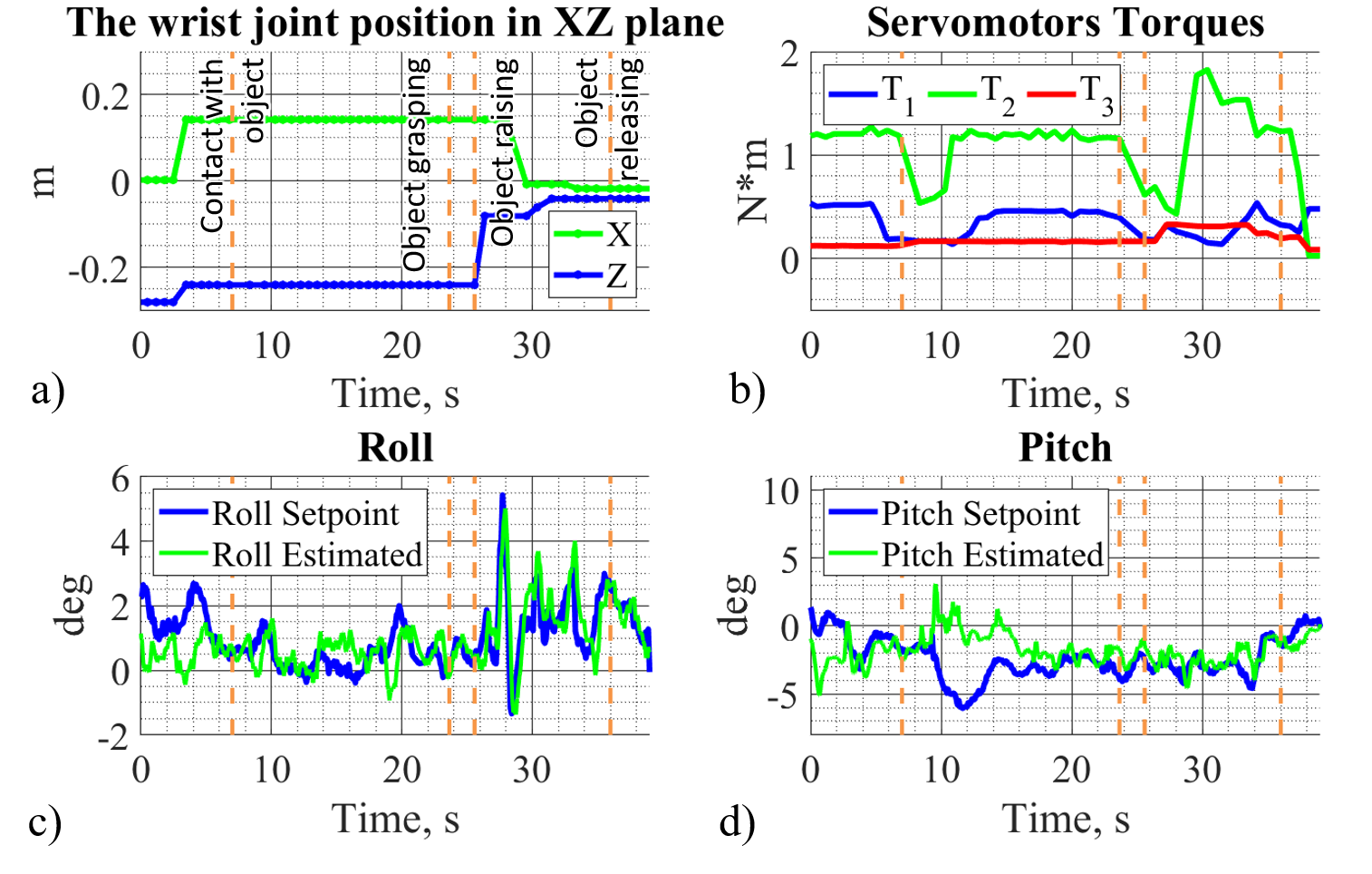}
\caption{Coordinates (a) and motor torques (b) of the manipulator, roll (c) and pitch (d) angles vs. time for the flight test with the VR-based teleoperation.} 
\label{fig:ExpRes2} 
\vspace{-0.5em}
\end{figure}

 Fig. \ref{fig:ExpRes2} shows the change of the position of the wrist joint, the torques of the servomotors of the manipulator, roll, and pitch angles of the UAV in time. The recorded time interval is shown between the moment when the robot hangs over the target position and the time when it is stabilized after releasing the object. Furthermore, the time-stamps of the key operations (contact with an object, object grasping, raising and releasing) are marked on this figure. The largest pitch deviation is 7.15 degrees (at 11.2 sec, Fig. \ref{fig:ExpRes2}(d)) that corresponds to the reliable contact of the manipulator with the object. Grasping the object and picking it after 2 seconds caused a significant fluctuation of the UAV attitude that affected on the roll angle (maximum standard roll deviation is of 2.61 degrees). After the object grasping, the drone was confined in movement in the direction along $Y$-axis. That is why its roll angle during this period (8 - 20 sec) deviated from the desired value, i.e. its set-point. The movement of the object in new position also increased the torque value in the elbow joint. The object release caused a sharp decrease of the robot mass (in comparison to the robot with the object) and the robot CoM moved quickly. The drone controller handled this situation, and the robot stabilized itself after 4 seconds. During aerial manipulation, standard deviations in pitch and roll directions were 2.03 and 0.83 degrees, respectively.

\par
To conclude, the teleoperation of aerial manipulation was successful. Tactile feedback from vibration motors of the glove effectively complements the visual information of the virtual environment, being a tangible confirmation of contact with the object. The main difficulties of the teleoperation were related to the data transfer from Unity to the manipulator controller. There were the cases when it manifested itself in delays in the execution of real movements of the manipulator. Throughout the experiments, a live telephone connection was used between the operator and the technical operator of UAV. It was necessary to perform the adjustment of the position of the robot in the perpendicular direction to the manipulation. In the future, it will not be necessary since we plan to add the function of the UAV control to the VR application.

\section{Conclusion}
We have proposed a VR-based teleoperation system for the UAV robotic manipulator. The flight tests of the developed robot showed a  stable UAV behavior while grasping of the target object.   The combined use of HTC VIVE trackers and designed glove demonstrated a robust data transmission to the Unity application. The average deviations in pitch and roll directions were 2.03 and 0.83 degrees, respectively.  The future work will be devoted to the development of the controller for decreasing deviations of the flying robot during the interaction with different objects and accurate robot position control in VR.

\addtolength{\textheight}{-12cm}  


\bibliographystyle{IEEEtran}
\bibliography{bib}

\end{document}